\documentclass[twoside,11pt]{article}

\usepackage[arxiv]{melba}

\usepackage{mwe} % to get dummy images, only for the example
\usepackage{amsmath,amsfonts}
\melbaid{YYYY:NNN}  % This is provided upon by the publishing editor
\doi{https://doi.org/10.59275/j.melba.2023-AAAA}
\melbaauthors{Rushcke, Carlsen, Hansen, Lindberg, Hindsholm, Norgaard \& Ladefoged}  % Note: this one is also used to set the pdf 'authors' metadata
\volume{2}
\firstpageno{1337}  % Communicated by the publishing editor
\melbayear{YYYY}  % The publication year
\datesubmitted{m1/yyyy}  % Date submitted to MELBA: mm/yyyy
\datepublished{m2/yyyy}  % Today's date: mm/yyyy

\ShortHeadings{Synthesizing Labeled Data to Identify Enlarged Ventricles}{}

\title{Guided Synthesis of Labeled Brain
MRI Data Using Latent Diffusion
Models for Segmentation of
Enlarged Ventricles}

\author{\firstname Tim \surname Ruschke \orcid{0009-0009-2444-6262} \email timruschke@gmail.com \\
    \addr Department of Computer Science, University of Copenhagen (UCPH), Denmark \\
    \addr DEPICT, Centre for AI and Medical Imaging, Rigshospitalet, Denmark
    \AND
    \firstname Jonathan Frederik \surname Carlsen \orcid{0000-0002-6724-7281} \email jonathan.frederik.carlsen@regionh.dk \\
    \addr Department of Radiology, Rigshospitalet, Denmark \\
    \addr DEPICT, Centre for AI and Medical Imaging, Rigshospitalet, Denmark \\
    \addr Department of Clinical Medicine, UCPH, Denmark
    \AND
    \firstname Adam Espe \surname Hansen \orcid{0000-0002-6457-1537} \email adam.espe.hansen@regionh.dk \\
    \addr Department of Radiology, Rigshospitalet, Denmark \\
    \addr DEPICT, Centre for AI and Medical Imaging, Rigshospitalet, Denmark \\
    \addr Department of Clinical Medicine, UCPH, Denmark
    \AND
    \firstname Ulrich \surname Lindberg \orcid{0000-0002-0004-6354} \email ulrich.lindberg@regionh.dk \\
    \addr Department of Clinical Physiology and Nuclear Medicine, Rigshospitalet, Denmark \\
    \addr DEPICT, Centre for AI and Medical Imaging, Rigshospitalet, Denmark
    \AND
    \firstname Amalie Monberg \surname Hindsholm \orcid{0000-0002-0049-0842} \email amalie.monberg.hindsholm@regionh.dk \\
    \addr Department of Clinical Physiology and Nuclear Medicine, Rigshospitalet, Denmark \\
    \addr DEPICT, Centre for AI and Medical Imaging, Rigshospitalet, Denmark
    \AND
    \firstname Martin \surname Norgaard \thanks{These authors have contributed equally to this work and share last authorship} \orcid{0000-0003-2131-5688} \email martin.noergaard@di.ku.dk \\
    \addr Department of Computer Science, University of Copenhagen (UCPH), Denmark \\
    \addr Molecular Imaging Branch, National Institute of Mental Health, United States
    \AND
    \firstname Claes Nøhr \surname Ladefoged \footnotemark[1] \orcid{0000-0002-3583-4190} \email claes.noehr.ladefoged@regionh.dk \\
    \addr Department of Clinical Physiology and Nuclear Medicine, Rigshospitalet, Denmark \\
    \addr DEPICT, Centre for AI and Medical Imaging, Rigshospitalet, Denmark \\
    \addr Department of Applied Mathematics and Computer Science, Technical University of Denmark, Denmark
}

\begin{document}

% top matter
\maketitle

% abstract
\begin{abstract}%   <- trailing '%' for backward compatibility of .sty file
	Deep learning models in medical contexts face challenges like data scarcity, inhomogeneity, and privacy concerns. Synthetic data can help, but often underperforms compared to real data. This study focuses on improving ventricular segmentation in brain Magnetic Resoncance Imaging (MRI) images, particularly for cases with enlarged ventricles, using synthetic data.
We employed two latent diffusion models (LDMs): a mask generator trained using 10,000 masks, and a corresponding SPADE image generator optimized using 6,881 scans to create an MRI conditioned on a 3D brain mask. Conditioning the mask generator on ventricular volume in combination with classifier-free guidance enabled the control of the ventricular volume distribution of the generated synthetic images. Next, the performance of the synthetic data was tested using three nnU-Net segmentation models trained on a real (N=1,000), augmented (N=1,512) and entirely synthetic data (N=1,000), respectively, where the synthetic data contains a more uniform distribution of ventricular volumes compared to the real data. The resulting models were tested on a completely independent hold-out dataset of patients with enlarged ventricles (N=42), with manual delineation of the ventricles used as ground truth performed by a trained neuroradiologist. The model trained on real data showed a mean absolute error (MAE) of $9.09\pm12.18$ mL in predicted ventricular volume, while the models trained on synthetic and augmented data showed MAEs of $7.52\pm4.81$ mL and $6.23\pm4.33$ mL, respectively. Both the synthetic and augmented model also outperformed the state-of-the-art model SynthSeg, which due to limited performance in cases of large ventricular volumes, showed an MAE of $7.73\pm12.12$ mL with a factor of 3 higher standard deviation. The model trained on augmented data showed the highest Dice score of $0.892\pm0.05$, slightly outperforming SynthSeg ($0.874\pm0.06$) and on par with the model trained on real data ($0.891\pm0.05$). The synthetic model ($0.872\pm0.06$) performed similar to SynthSeg. In summary, we provide evidence that guided synthesis of labeled brain MRI data using LDMs improves the segmentation of enlarged ventricles and outperforms existing state-of-the-art segmentation models. 
\end{abstract}

% keywords

% Introduction (or first section)
\section{Introduction}
	The success of machine learning largely depends on the quantity and quality of available training data. As models grow increasingly large, they require immense datasets, which has culminated in some of the most successful models trained on so many instances of data they need not see the same sample twice (\cite{brown2020language_gpt3}). While such abundance can be found in natural images or natural language processing tasks, this wealth of data is not universally accessible. In medical imaging, data scarcity, variability, and privacy constraints present significant challenges to building similarly large models.

Patients with normal pressure hydrocephalus (NPH) exemplify this issue, displaying an abnormal buildup of cerebrospinal fluid in the brain's ventricle system, consequently leading to unique and abnormally enlarged ventricles. Contemporary state-of-the-art models for brain segmentation of magnetic resonance imaging (MRI) data (\cite{billot_robust_synthseg_2023}) struggle to accurately segment the enlarged ventricle systems of NPH patients. Furthermore, obtaining a large training set for these patients is difficult due to their uniqueness and the diverse nature of the disease.

Generating synthetic data offers a promising alternative to the prohibitively costly process of collecting real data. Modern image generation models, such as diffusion models, initially seem to provide limitless, highly customizable data with perceptual quality nearly indistinguishable from real images (\cite{podell2023sdxl}). However, extensive research has shown that purely synthetic training data cannot match the effectiveness of real training data (\cite{marwood2023diversity_diffusion,Fernandez_2022_trained_with_fully_synthetic_data,Fernandez_2024_SyntheticData}). Instead, synthetic data is best used to supplement real data to enhance model performance (\cite{azizi2023synthetic_improves_imagnet}). 

In the field of MRI, several studies (\cite{pinaya2022brain_image_generation_w_LDM, han2023medgen3d}) have demonstrated the remarkable perceptual quality of synthetic images generated using diffusion models. Although these synthetic images still do not match the quality of real data, \cite{Fernandez_2022_trained_with_fully_synthetic_data} made significant progress in narrowing the performance gap between real and synthetic training data for brain MRI segmentation models. Additionally, conditioning the models on scalar values can produce increasingly realistic and morphology-preserving images. For example, \cite{pinaya2022brain_image_generation_w_LDM} trained a latent diffusion model (LDM) on a large sample of 3D T1-weighted MRI data (N=31,740) from the UK Biobank dataset (\cite{Sudlow2015_UK_Biobank}) while conditioning on several parameters such as ventricular volume, thus being able to generate synthetic data with varying ventricle sizes. Achieving remarkable visual quality, \cite{pinaya2022brain_image_generation_w_LDM} used their model to create an open large synthetic dataset consisting of 100,000 synthetic 3D MRI images (LDM100k dataset). 

In this study, we extend prior work by presenting two LDMs: a mask generator and an image generator. Combined, these models create fully synthetic labeled segmentation data for brain MRI. By conditioning our mask generator on ventricular volume, we enable control over the distribution in the generated data. We generate synthetic images and labels with out-of-distribution (OOD) ventricle sizes using our LDMs to address data scarcity for patients with enlarged ventricles. Finally, we train a ventricle segmentation model using this synthetic data and compare the performance against state-of-the-art models, using a completely independent test set consisting of NPH patients with enlarged ventricles.

\setcounter{footnote}{0}
Code to reproduce this work is publicly available on GitHub.\footnote{See \url{https://github.com/vaynonym/diffusion_abnormal_brain_anatomy}}

%%%%%%%%%%%%%%%%%%%%%%%%%%%%%%%%%%%%%%%%%%%%%%%%%%%%%%%%%%%%%%%%%%%%%%%%%%%
% Related works
%%%%%%%%%%%%%%%%%%%%%%%%%%%%%%%%%%%%%%%%%%%%%%%%%%%%%%%%%%%%%%%%%%%%%%%%%%%
% Make sure to put your work into context and include apporpriate citations.
% We do not have limits on citation counts.

% A methodological, model, or similar section often comes here.
\section{Methods}
	\subsection{Data}
 \newcommand{\Dldm}{LDM100k Dataset}
\newcommand{\Dflair}{MS-FLAIR Dataset}
		\subsubsection{\Dldm}\label{DLM_dataset_definition}

%With real medical data being subject to restrictions in where and how it may be used, the use of synthetic data offers several tempting advantages. Furthermore, as synthetic data has reached a level of quality where it can be used to train models successfully on some downstream tasks \cite{Fernandez_2022_trained_with_fully_synthetic_data}, it can be a viable option to explore further. 

\cite{pinaya2022brain_image_generation_w_LDM} generated a publicly available\footnote{Data can be downloaded from \url{https://academictorrents.com/details/63aeb864bbe2115ded0aa0d7d36334c026f0660b}} large synthetic dataset of 100,000 3D T1-weighted brain MRI images using an LDM trained on UK biobank data (\cite{Sudlow2015_UK_Biobank}) with isotropic $1$ mm$^3$ voxel size and a resolution of $160$ x $224$ x $160$. In this work, to reduce computational constraints, we first downsampled the data to a spacing of $1.67$ x $1.75$ x $1.67$ $mm^3$ and center-cropped to a resolution of $96$ x $128$ x $96$. We also only used the first 10,000 images in the data to keep the size in line with our other dataset (\Dflair{}), assuming a valid representation of the full dataset. The main purpose of this dataset is two-fold, 1) serving as the training data for our mask diffusion model due to its balanced spread of ventricular volumes, and 2) also being used for transfer learning. To generate segmentation labels for the data, we used the SynthSeg model (\cite{billot_robust_synthseg_2023}). In addition to the segmentation, the model also estimates the size of the volumes for different parts of the brain. To obtain the conditioning value $c$, we divided the ventricular volume by the intracranial volume to get the relative size of the ventricles. We subsequently normalized these values between 0 and 1 over the subset of the data, and the outcome was then parsed into the models for conditioning. Lastly, we used a randomized 80-20 schema, splitting the data into training and hold-out validation datasets.

\subsubsection{MS-FLAIR Dataset}

The dataset consists of T2-weighted 3D FLAIR images acquired at the Copenhagen University Hospital - Rigshospitalet, Copenhagen, Denmark. The data comprises 2,927 patients with multiple sclerosis (MS) who were scanned multiple times as part of their treatment, resulting in a total of 6,881 scans. A more thorough description of the data is available in \cite{hindsholm_scanner_2023}. All patient-specific data was handled in compliance with the Danish data protection agency act no. 502. The collection of the retrospective dataset was approved by the National Committee on Health Research Ethics (protocol number 2117506). 

We performed the same steps as for the LDM100k data to obtain segmentation masks and corresponding volumetric estimates for each MRI. Since the LDM100k dataset was spatially aligned, we further cropped the MS-FLAIR dataset placing the brain in the center of the image using the segmentation masks, before downsampling to the same spacing and cropping to the same image size of $96$ x $128$ x $96$.

\begin{figure}[t!]
    \centering
    \includegraphics[width=0.98\linewidth]{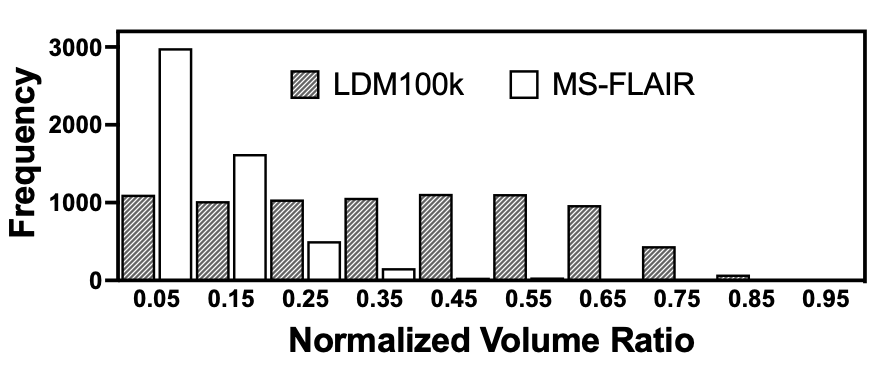}
    \caption{Histogram of the distribution of normalized ventricular volumes for the LDM100k data (N=10,000) and the MS-FLAIR data (N=6,881). The x-axis displays the normalized volume ratio $c$, which is estimated as the ratio of ventricular volume to intracranial volume to get the relative size of the ventricles, and subsequently normalized over the data to the range between 0 and 1.}
    \label{fig:dataset_ventricular_volume_comparison}
\end{figure}

A comparison of the distribution of images with respect to ventricular volume can be seen in \autoref{fig:dataset_ventricular_volume_comparison}. It is evident that the \Dflair{} is much less uniformly distributed with respect to ventricular volume than the \Dldm. 

\subsection{Diffusion Models}

Diffusion models are generative models that can be used to generate synthetic images by iteratively denoising a noisy sample, starting from simple gaussian noise. This is called the backward process, modeling the reverse of the forward process which iteratively adds noise to an image until reaching complete noise. Given enough time steps in the procedure, there is a mathematical justification (\cite{original_diffusion_paper}) that if the forward process follows a gaussian distribution, so does the backward process. The gaussian distribution modeling the backward process is implemented by means of a neural network, indicated by parameters $\theta$. As each step is independent from the previous one, leveraging the ensuing markov property results in the entire backward trajectory given by 

\begin{equation}
    p_\theta(\textbf{x}_{0:T}) = p(\textbf{x}_T) \prod_{t=1}^T p_\theta(\textbf{x}_{t-1}| \textbf{x}_t)
\end{equation}

where $T$ is the number of timesteps,     $p(\textbf{x}_T) = \mathcal{N}(\textbf{x}_T; \mathbf{0}, \mathbf{I})$, $x_0$ a sample from the data distribution, and $    p_\theta(\textbf{x}_{t-1}| \textbf{x}_t) = \mathcal{N}(\textbf{x}_{t-1}; \mathbf{\mu_\theta}(\textbf{x}_t, t),\mathbf{\Sigma_\theta}(\textbf{x}_t, t))$. The forward process is modeled analogously but with a known distribution based on a specific noise schedule $\beta_t$.

For training, \cite{original_diffusion_paper} derived the evidence lower bound (ELBO) for the log likelihood (intractable to calculate directly), which has since been simplified and improved by \cite{DDPM}. After making a few minor assumptions about the model, rewriting the objective in terms of predicting the noise instead of the the target image, and simplifying as much as possible, the final objective function is given by 

\begin{equation}
    \frac{\beta_t^2}{2\sigma_t^2 \alpha_t (1-\bar{\alpha}_t)}  ||\epsilon - \epsilon_\theta(\sqrt{\Bar{\alpha}_t} x_0 + \sqrt{1- \bar{\alpha}_t}\epsilon, t)||^2
\end{equation}

which compares the noise added to the
original image with the noise predicted by the neural network using mean squared error (MSE),
scaled by a factor depending on the choice of noise schedule $\beta_t$ (from which $\alpha_t$ and $\bar{\alpha_t}$ are derived, see \cite{DDPM}). The training algorithm (see \cite{DDPM}) is derived from the expectation over the data and over the time of the learning objective, where we then simply sample from the training data, sample a timestep $t \sim [1, T]$ uniformly at random, and then adjust the model by the gradient according to the loss calculated.

\subsection{Conditioning Diffusion Models}\label{Background: Conditioning}

The most simple case of conditioning a given distribution is to simply add a conditioning input $c$ at each step as an input to the model, as given by

\begin{equation}
    p(\textbf{x}_{0:T} | c) = p(\textbf{x}_t) \prod^T_{t=1} p_\theta(\textbf{x}_{t-1}| \textbf{x}_t, c )
\end{equation}

which simply corresponds to $\epsilon_\theta(\textbf{x}_t, t, c)$ getting an additional parameter. In practice, this can take the form of concatenating the conditioning information with the input image or using an attention mechanism (\cite{rombach2022_highresolution_LDM_synthesis,attention_is_all_you_need}). 

Guidance (\cite{dhariwal2021diffusionModelsBeatGANs_classifier_guidance}) has been suggested as a means to amplify the effect of conditioning. Guidance uses the gradient of a classifier (corresponding to $p(c | \textbf{x}_t)$) to influence the prediction step when sampling. As this mechanism requires a separately trained classifier, classifier-free guidance (\cite{ho2022ClassifierFreeGuidance}) is a convenient way to forgo the classifier, instead training both a conditional and unconditional model (which can even be the same model training for both cases) and combining the predictions with a weight $G$:

\begin{equation*}\label{Classifier-Free Guidance}
    \epsilon_\theta = G \epsilon_{p_\theta(\textbf{x}|c)} + (1 - G) \epsilon_{p_\theta(\textbf{x})} 
\end{equation*}

Due to the scaling by G in both terms, similarities in the cases are independent from G while differences will be emphasized in favor of the conditional case as G increases beyond 1.

\subsection{Mask Generator Latent Diffusion Model}

In this work, the first step in our model development pipeline to generate synthetic labeled data is to generate masks. For this, we used a diffusion model, specifically an LDM  (\cite{rombach2022_highresolution_LDM_synthesis}) due to the large computational requirements imposed by working in the 3D image space. The autoencoder architecture and hyperparameters were based on \cite{pinaya2022brain_image_generation_w_LDM}, but given the different target of mask reconstruction instead of images, we made several changes based on an informal hyperparameter search. First, our reconstruction loss consisted of a cross-entropy loss. As labels cannot be trivially used with common perceptual losses and we found no benefit in using an adverserial loss, we used no other additional loss to train the model except for KL-regularization to a normal distribution with a weight of $1\mathrm{e}{-6}$. Next, instead of feeding the labels directly to the model, we used an embedding layer, trained end-to-end, before the first layer, with a feature dimension of 64. We trained the autoencoder for 40 epochs using the \Dldm. For the diffusion model, we followed the basic denoising model and training algorithm from \cite{DDPM}. The backbone of the noise-prediction model was a U-Net with an additional attention mechanism closely resembling \cite{pinaya2022brain_image_generation_w_LDM}. We used conditioning on the normalized ventricular volume ratio $c \in [0, 1]$ (see \autoref{DLM_dataset_definition}) via attention. Further, we used classifier-free guidance (\cite{ho2022ClassifierFreeGuidance}) by training the model on the conditional and unconditional case at the same time, adding an additional binary conditioning parameter and randomizing $c$ in the unconditional case. The extra parameter was needed to ensure $c$ maintained a smooth distribution which the model can learn more easily. During training, there was a 20\% chance for any dataset entry to be transformed to the unconditional case. The diffusion model was trained for 5 epochs. The low number of epochs was motivated by \cite{dar2024unconditional_memorize_patient_data}, who found that longer training times are associated with reproducing patient data and thus less diversity. For more details, please see \autoref{fig:mask_generator_overview} in the appendix.

\subsection{Image Generator Latent Diffusion Model}

\begin{figure}[t!]
    \centering
    \includegraphics[width=0.97\linewidth]{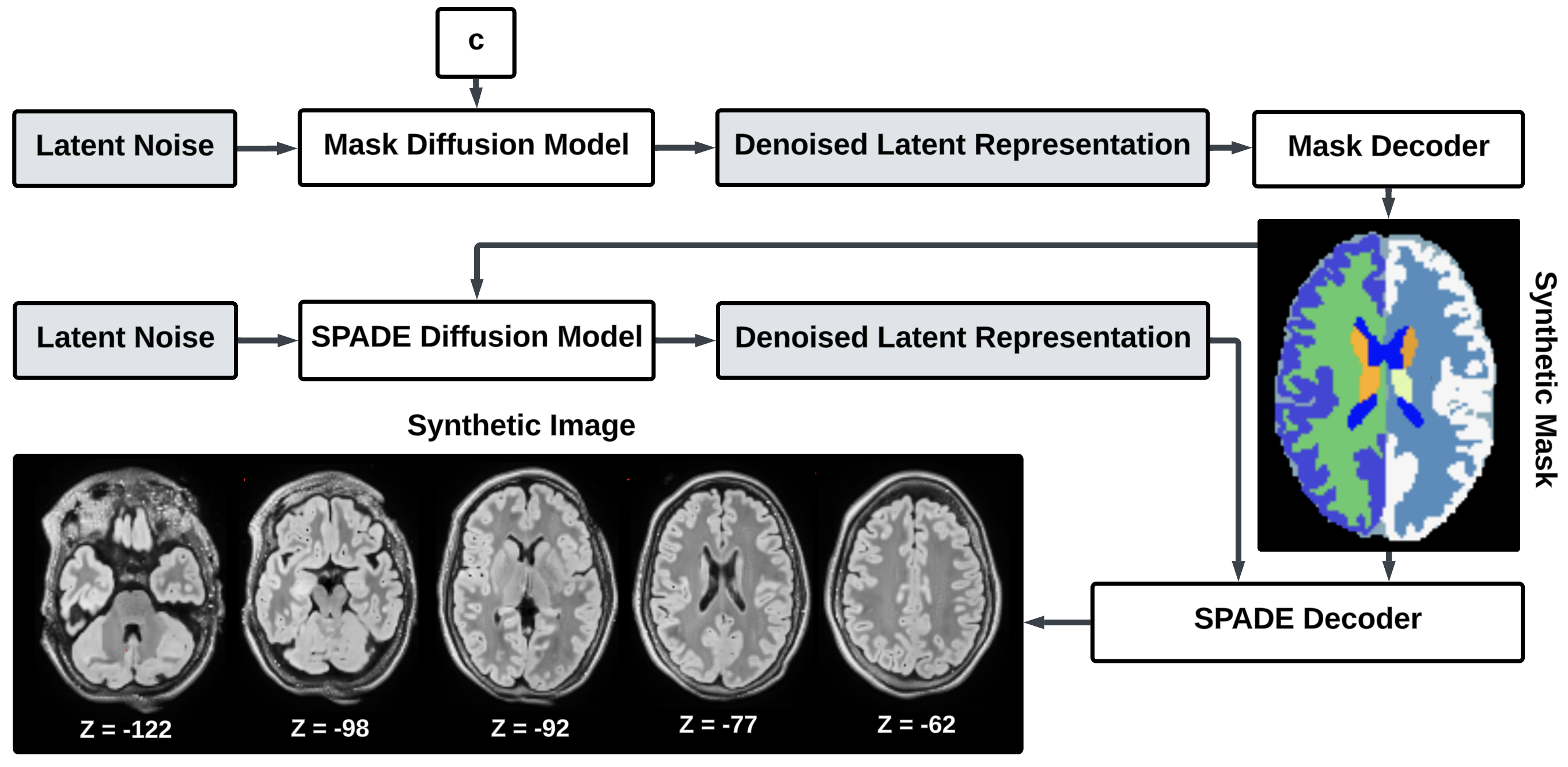}
    \caption{Overview of the pipeline used to generate synthetic labeled data. Latent diffusion models (DM) are employed in two stages to produce synthetic brain images and corresponding masks. The pipeline begins with a noise input passed through a Mask DM that can be conditioned on the parameter $c$ reflecting the normalized ventricular volume ratio, followed by denoising to produce a latent representation. A Mask Decoder transforms this latent representation into a synthetic segmentation mask. Parallel to this, a SPADE DM, conditioned on the synthetic mask, generates denoised latent representations which are decoded to create a synthetic brain image. }
    \label{fig:data_generation_inference_architecture}
\end{figure}

Now having defined a mask generation LDM (previous subsection), we used the output of that model to condition the image generation (Figure \ref{fig:data_generation_inference_architecture}), following the same general 3D LDM approach.
Specifically, we used a KL-autoencoder similar to the Mask Generator case, but the decoder used the segmentation mask like a SPADE model (\cite{park2019_Spade}) to enable precise alignment of images with the mask. We used a similar loss combination as in \cite{pinaya2022brain_image_generation_w_LDM}, using a combination of perceptual loss, adverserial loss, pixel-based L1 loss, and KL-loss as in the mask generator case. For the adverserial loss, we used a patch-based discriminator based on \cite{wang2018highresolution_patch_discriminator_basis}. We trained the autoencoder on the \Dldm{} for 20 epochs first to utilize the large and diverse dataset, and then fine-tuned the autoencoder for another 10 epochs on the \Dflair, since the test set used for the downstream segmentation task was acquired using the T2-weighted FLAIR sequence. The diffusion model closely matched the mask generator model architecture. Like the decoder of the autoencoder, the upsampling part of the U-Net backbone also employed SPADE connections. We trained the diffusion model for 20 epochs using the \Dldm{} and fine-tuned for 5 epochs on the \Dflair. 
For an overview of the general training architecture, please see \autoref{fig:Spade_model_architecture} in the appendix.

\section{Experiments}
\newcommand{\Dsyn}{$D_{syn}$}
\newcommand{\Dreal}{$D_{real}$}
\newcommand{\Daug}{$D_{aug}$}

\newcommand{\testsettwo}{NPH-FLAIR$_{test}$}

\subsection{Generating large ventricles with guidance and conditioned on out of distribution values}\label{Experiment: Generating large ventricles with guidance and out of distribution Values}

We trained our model on the \Dldm{} and compared the generated distribution with the ground truth values (masks generated using SynthSeg (\cite{BILLOT2023102789})) from the validation set. Model values were determined by generating synthetic masks in equal number with the same $c$ values as present in the validation set. With matching $c$ values, we should expect similar mean volumes in the generated data, assuming the model generalizes well. We repeated this for $G \in \{1, 2, 3, 4, 5\}$, with $G=1.0$ corresponding to the conditioned case without guidance. Due to the long generation time of an LDM, we used a DDIM scheduler (\cite{song2022DDIM}) with 50 inference timesteps to speed up the process. For values of $c > 1.0$, we sampled 50 images per bucket, each prompted with the middle point of the bucket (1.05, ...). 

\begin{figure}[t!]\includegraphics[width=0.98\linewidth]{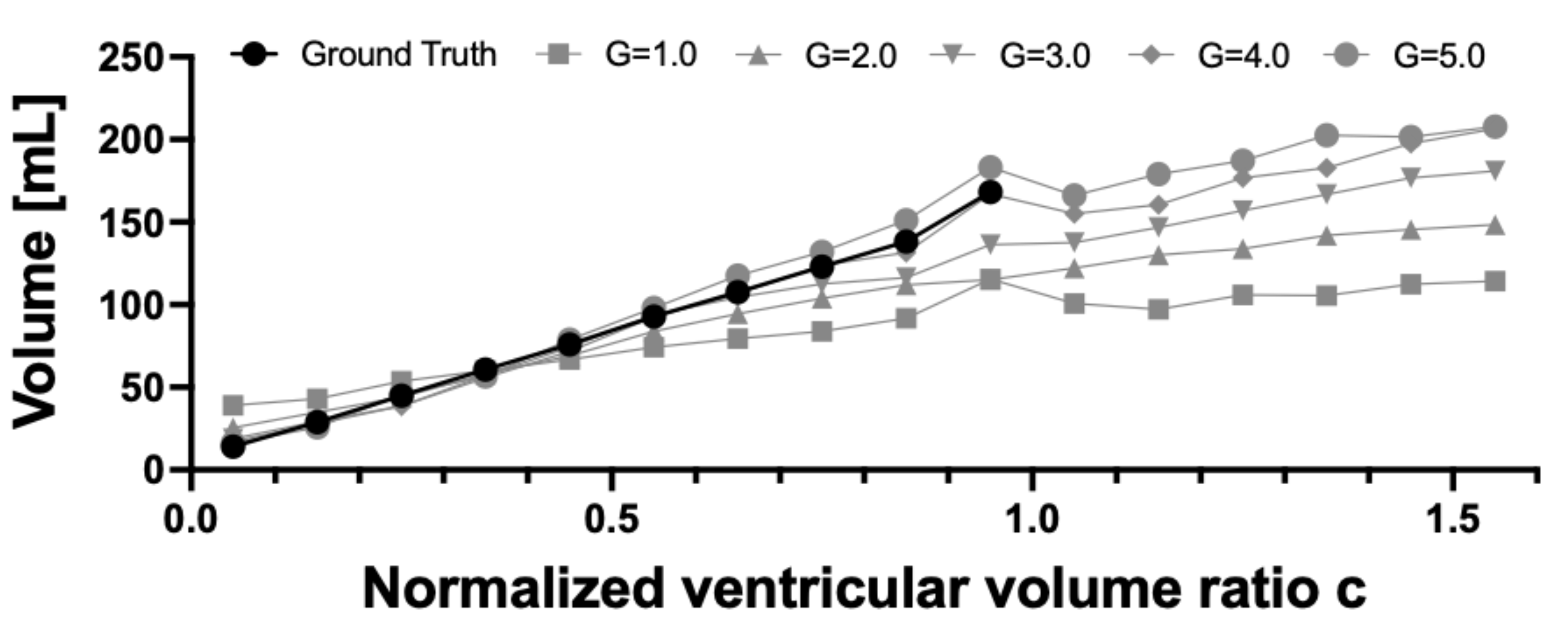}
    \caption{Overview of the effect of conditioning and guidance on ventricular volume generation. Specifically, this figure displays the mean ventricular volume (in mL) as a function of the normalized ventricular volume ratio $c$, comparing the ground truth ventricular volumes (black circles) with synthetic data generated using different guidance factors $G \in \{1.0, 2.0, 3.0, 4.0, 5.0\}$. The ground truth values are derived from the validation set using SynthSeg-generated masks. Synthetic masks were generated with similar $c$ values, and the mean ventricular volumes were compared. We also prompt the model with $c > 1.0$ (values not encountered during training) to investigate how well the model can extrapolate to larger ventricular sizes.}
    \label{fig:LDM100k_guidance_ventricular_volume}
\end{figure}

We start by studying the behaviour of the generated distribution of ventricular volumes when varying the conditioning and guidance parameters, reported in Figure \ref{fig:LDM100k_guidance_ventricular_volume}. In Figure \ref{fig:LDM100k_guidance_ventricular_volume}, the ground truth curve follows a roughly linear shape, which is to be expected given that $c$ is calculated from the ventricular volume of the masks.
While the conditional case $G=1.0$ follows an approximately linear curve, the slope is negatively biased, over-shooting for small values of $c$ and under-shooting as $c$ increases. Even for out of distribution values, the curve never reaches the peak ground truth value. However, as the guidance factor $G$ increases, a linear relationship between $G$ and the slope of the curve is observed. The guidance factor $G=4.0$ roughly matches the ground truth distribution, while $G=5.0$ produces greater mean ventricular volumes than the ground truth reference. This demonstrates that the model can be prompted to generate images with not only large ventricular volumes, but even larger than encountered in the training distribution, with the possibility to change the parameters as desired at inference time without retraining the model. 

%During additional exploratory analyses, we found that training for more epochs resulted in a closer match of distribution even at lower guidance values. However, this comes with two shortcomings. First, we noticed an increased rate of degeneration at inference. Second, longer training times are generally associated with less diversity. On the other hand, training the mask generator \Dflair{} massively undershoots the ground truth distribution due to the extreme sparsity at medium and higher $c$ values, suggesting the need for some base diversity in the dataset to accurately learn the distribution. Graphs for both can be seen in the addendum.

%For an analysis of how well the generated distribution matched the training distribution in terms of size of distinct anatomical regions, please refer to the addendum. \todo{reference}

\subsection{Image generation conditioned on synthetic masks}

\begin{figure}[t!]
\centering
    \centering
    %\textbf{Ventricular Volume}\par\medskip
    \includegraphics[width=\linewidth]{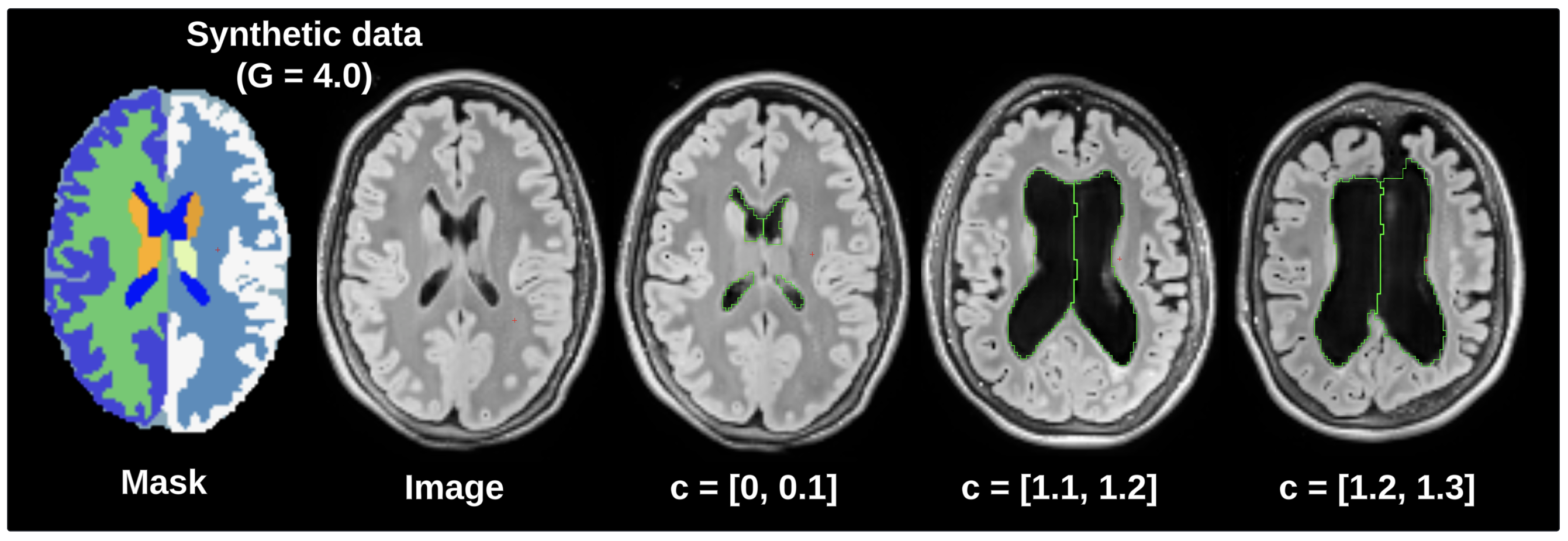}
    \caption{Example case of synthetic brain images/masks generated using a guidance factor $G=4.0$ and varying conditioning parameters $c$. On the left, the synthetic segmentation mask is displayed, followed by the corresponding synthetic brain image. The right-most three images show variations in ventricular size for increasing $c$ values, with the ventricular label outline highlighted in green. As $c$ increases, the ventricular size expands accordingly, reflecting the expected changes based on the conditioning parameter. Degeneration of the synthetic data is observed for increasing $c$ values (right).}
    \label{fig:LDM100k_guidance_ventricular_volume_MRI}
\end{figure}

Figure \ref{fig:LDM100k_guidance_ventricular_volume_MRI} shows representative samples of synthetic images, when varying the conditioning parameter $c$ at $G=4.0$. We generally observed degeneration of the synthetic data (Figure \ref{fig:LDM100k_guidance_ventricular_volume_MRI}, right) for $c>1.4$ or $G>4.0$, but precise thresholds varied across models. The transition took the form of a gradual increase in probability of degeneration, as we rarely observed coherent samples for large $c$ and $G$ and very infrequently for out of distribution values that were still somewhat close to the training distribution, like $c=1.25$, $G=4.0$. However, it is non-trivial to develop a reliable mechanism for detecting degenerate masks to quantitatively analyze the rate, a topic which is left for future research.

We report quantitative metrics for assessing the image quality of the generated images for the diffusion model fine-tuned on the \Dflair{} compared against the hold-out validation set. The metrics serve mostly for comparison in future work, as they can be difficult to interpret across datasets. We used the Fréchet Inception Distance (FID) (\cite{heusel2018gans_fid}) as a measure of realism, where a small value indicates that the distribution of the generated images are similar to the distribution of the real images. We also measured the diversity of the generated images by computing the Structural Similarity Index Measure (SSIM) and the Multi Scale SSIM (MS-SSIM) between 500 pairs of synthetic images, with low values indicating high diversity. The model had an FID of 0.34, a pairwise SSIM of 0.31, and a pairwise MS-SSIM of 0.45. 

In addition to the quantitative metrics, we also performed a qualitative analysis by having an experienced neuroradiologist assess the realism and quality of a smaller subset of the generated images. For this, we generated thirteen synthetic images with a guidance value of 4.0 and $c\in[0, 1.3]$ in intervals of 0.1, and further 10 with guidance 1.0 and $c\in[0, 1.0]$. All generated images used a DDPM scheduler with 1,000 timesteps. The generated images were visually inspected by a neuroradiologist, reporting that the images were generally realistic, particularly noting the correct displacement of surrounding tissue caused by enlarged ventricles. However, larger guidance values were generally associated with less realistic images.

\subsection{Can synthetic data improve segmentation performance on enlarged ventricles?}\label{Downstream_Tasks}

\begin{figure}[t!]
    \includegraphics[width=0.9\linewidth]{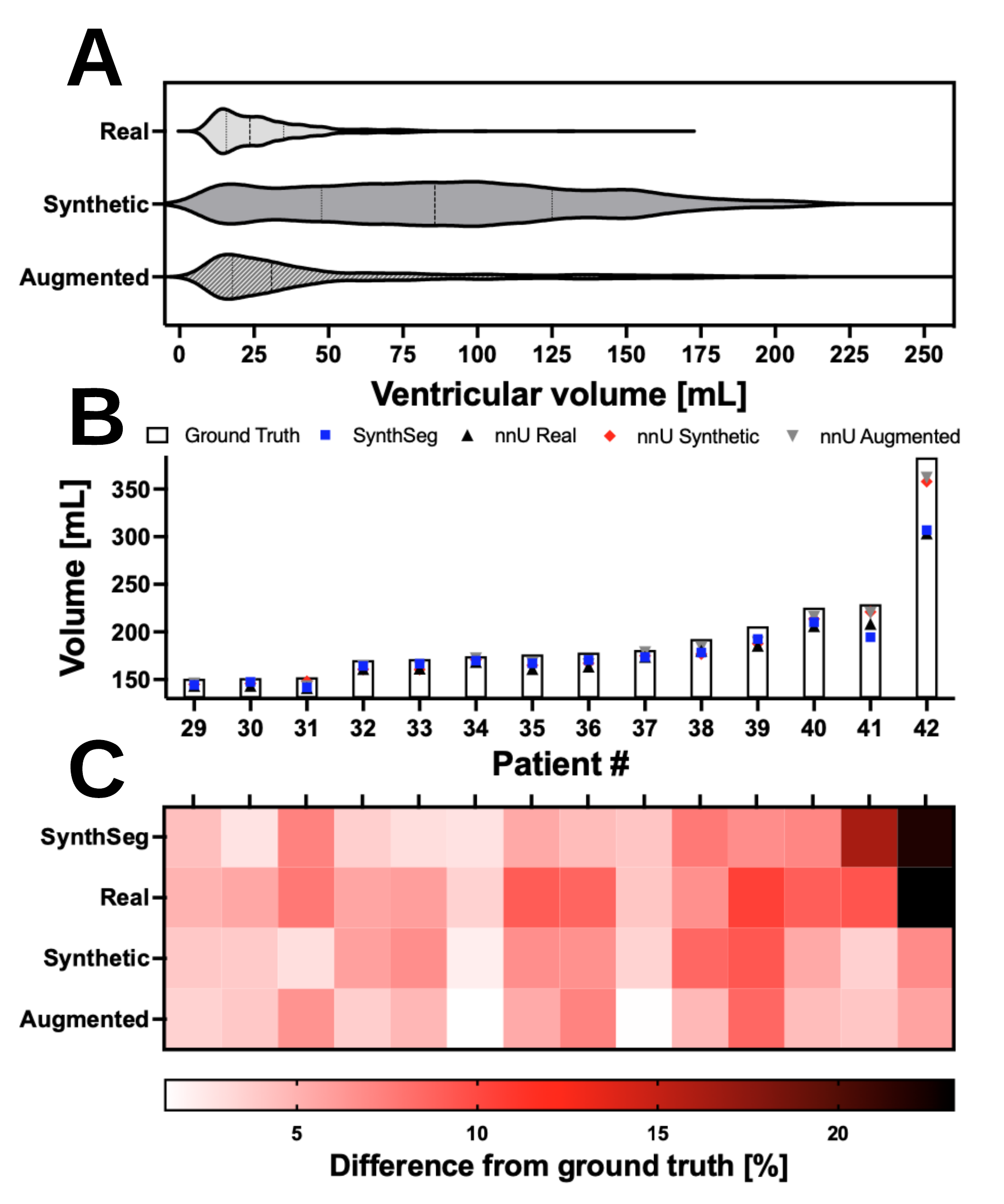}
    \caption{\textbf{(A)} Violin plots of ventricular volume for each of the three training datasets using real, synthetic, and augmented (synthetic+real) \textbf{(B)} volume of predicted ventricle mask for a subset of the \testsettwo{} (N=42) with ventricular volumes above 150 mL, sorted by ground truth volume, \textbf{(C)} percent difference from ground truth and the predicted ventricular volume across the 4 models for the same subset of data as in (B).}
    \label{fig:data_dist_prediction}
\end{figure}

To test whether synthetic data can improve the segmentation performance on enlarged ventricles, we created three datasets, \Dreal, \Dsyn, and \Daug. \Dreal{} consisting of 1,000 real images with labels from SynthSeg randomly sampled from the previously used FLAIR training data (Figure \ref{fig:data_dist_prediction}A). \Dsyn{} consisted of paired synthetic labels and synthetic images, generated using the mask generator and image generator SPADE model. The data was systematically generated to have a more uniform distribution of ventricular volumes compared to the real data, including out of distribution values. Specifically, we created 600 images using a guidance value of $4.0$ and a conditioning value  $c \in [0, 1.3]$ with intervals of size $0.1$ containing the same number of images. The exact value of $c$ in an interval was sampled uniformly at random. Motivated by enhancing diversity as well as robustness of the images, we generated another 400 images using the same procedure but with guidance $G=1.0$ and $c \in [0, 1.0]$. Lastly, \Daug{} consisted of a subset of \Dsyn{} (200 with $G=1.0$, 312 with $G=4.0$) as well as the entirety of \Dreal{}, comprising 1,512 samples in total. The roughly 2:1 ratio of real to synthetic data was a conservative estimate motivated by previous research, showing that too much synthetic data can be detrimental (\cite{azizi2023synthetic_improves_imagnet,Fernandez_2024_SyntheticData}).

In order to investigate the impact of the three training datasets (real, synthetic, and augmented), we subsequently trained several models for a downstream segmentation task. Specifically, we used the original nnU-Netv2 segmentation model (\cite{isensee_nnu-net_2021}) with no further modifications to the model or training other than the data, as this model has been shown to achieve state-of-the-art performance on many segmentation tasks. We used the nnU-Net training procedure algorithm out of the box, but only used the 3D full resolution configuration as our resolution was already quite small. We trained the model with mirroring removed from the data augmentation pipeline because otherwise, the model would not have been able to properly distinguish between left and right anatomical structures.

To evaluate the performance, we ran inference of our three models on \testsettwo{}, which consisted of 42 patients with enlarged ventricles imaged with the T2-weighted FLAIR sequence. The ground truth ventricle labels were manually delineated by an expert. A large proportion of these images used a 2D T2w-FLAIR acquisition instead of 3D, and thus differ from the training data of both the generator models as well as the nnU-Net models.
Finally, we also compared the trained nnU-Net models to the performance of SynthSeg (\cite{BILLOT2023102789}).

\newcommand{\NNUreal}{nnU$_{real}$}
\newcommand{\NNUaug}{nnU$_{aug}$}
\newcommand{\NNUsyn}{nnU$_{syn}$}

\begin{table}[ht]
\centering
\resizebox{\linewidth}{!}{%
\begin{tabular}{|c|c|c|c|c|}
\hline
\textbf{Model Type} & \textbf{Dice} & \textbf{IoU} & \textbf{Volume MAE} & \textbf{Volume MSE} \\ \hline
\NNUreal & $0.891 \pm 0.046$ & $\mathbf{0.807 \pm 0.069}$ & $9.09 \pm 12.18$ & $231.02 \pm 959.74$ \\ 
\NNUsyn & $0.872 \pm 0.058$ & $0.778 \pm 0.085$ & $7.52 \pm 4.81$ & $79.74 \pm 110.73$ \\ 
\NNUaug & $\mathbf{0.892 \pm 0.047}$ & $0.807 \pm 0.072$ & $\mathbf{6.23 \pm 4.33}$ & $\mathbf{57.50 \pm 83.24}$ \\ \hline
SynthSeg & $0.874 \pm 0.056$ & $0.781 \pm 0.084$ & $7.73 \pm 12.12$ & $208.03 \pm 893.28$ \\ \hline
\end{tabular}
}
\caption{Segmentation performance metrics and errors in predicted ventricular volumes for each model on the \testsettwo{} data. Segmentation accuracy is measured using Dice and Intersection over Union (IoU) scores. The volume prediction errors are reported as Mean Absolute Error (MAE) and Mean Squared Error (MSE), with values reported in mL.}
\label{tab:testset_2_binary_metrics}
\end{table}

%In Table \autoref{tab:testset_2_binary_metrics}, the segmentation performance is shown for each model. On Dice and IoU, \NNUaug{} and \NNUsyn{} outperform \NNUreal{} by a small margin while also displaying less standard deviation, indicating stronger consistency. All models performed slightly better compared to SynthSeg, but ultimately achieved very similar results. While the segmentation metrics of Dice and IoU provide a good measure of how well the voxels of the generated segmentation align with the ground truth, they only indirectly measure whether the ventricular size was predicted correctly, and particularly at this low resolution can be quite sensitive to voxel shifts. As such, we additionally compared the ventricular volume between predicted and ground truth masks. As we are interested in its performance on outliers with enlarged ventricles, which we assume to be more difficult to segment correctly, we measured not only mean absolute error (MAE) but also mean squared error (MSE). \NNUaug{} performed the best in terms of MAE, but displayed higher standard deviation and even a slightly higher MSE than \NNUsyn. Meanwhile, both \NNUreal{} and SynthSeg displayed much greater standard deviation than the two variants with synthetic data, and correspondingly a MSE almost four times as high. Particularly interesting is that while SynthSeg performed far better than \NNUreal{} in MAE, it performed almost equally bad at MSE, indicating the presence of some significant outliers for the SynthSeg model.

In \autoref{tab:testset_2_binary_metrics}, the segmentation performance is shown for each model.  On Dice and IoU, \NNUaug{} and \NNUreal{} perform equally well and both outperform \NNUsyn{} and SynthSeg by a small margin while also displaying less standard deviation. SynthSeg performs similarly to \NNUsyn{}. While the segmentation metrics of Dice and IoU provide a good measure of how well the voxels of the generated segmentation align with the ground truth, they only indirectly measure whether the ventricular size was predicted correctly, and particularly at this low resolution can be quite sensitive to voxel shifts. As such, we additionally compared the ventricular volume between predicted and ground truth masks. As we are interested in its performance on outliers with enlarged ventricles, which we assume to be more difficult to segment correctly, we measured not only mean absolute error (MAE) but also mean squared error (MSE). Both models using synthetic data (\NNUsyn{} and \NNUaug{}) achieved the lowest MAE and MSE. While SynthSeg was close to \NNUsyn{} in MAE, it showed a factor of 3 higher standard deviation, and also high MSE indicating the presence of outliers. \NNUreal{} showed lower MAE compared to all other models, and also high MSE similar to SynthSeg. 

Figure \ref{fig:data_dist_prediction}B provides an overview of predicted versus ground truth volume estimates for each model for a subset of larger ventricular volumes. In most cases, especially for patients with ventricular volumes up to 150 mL, all four models performed similarly, which can be explained by each model having ventricles of those sizes as part of the training data. However, as seen in Figure \ref{fig:data_dist_prediction}C, several patients with ventricle volumes above 200 mL resulted in poor performance for \NNUreal{} and SynthSeg, while \NNUsyn{} and \NNUaug{} retained consistent performance. \autoref{fig:side_by_side_patient_42} in the appendix shows a side-by-side view of each model's segmentation for the patient with the highest ventricles. The difference in performance between the models can be explained by the two latter models having seen synthetic samples with ventricular volumes above 200 mL during training.

Lastly, all models consistently showed undershooting compared to the ground truth volume size, with the exception of the patient with the smallest volume. Speculatively, this could be due to a difference in labeling style between SynthSeg and the human expert.

\section{Discussion}

The prospect of training a model for a given task solely on synthetic data is a tempting one. Between a potential abundance of training data and reduced (albeit not \textit{removed} (\cite{dar2024unconditional_memorize_patient_data})) concern for its privacy, this approach appears to address several pervasive problems with machine learning in a medical context. 

In this work, building on previous successful 3D synthesis of brain MRI (\cite{pinaya2022brain_image_generation_w_LDM}), we developed a separate LDM mask generator and SPADE LDM image generator to create paired synthetic images and labels. While other approaches cast the mask reconstruction as an image reconstruction problem (\cite{Fernandez_2022_trained_with_fully_synthetic_data}), we employed a more simple approach where the mask was encoded via an end-to-end trained embedding and used a simple cross-entropy loss commonly used in segmentation problems. Our own hyperparameter search on this problem was only cursory, and given the label in our architecture dictated the semantics and shape of the final image, the role in final image diversity is likely very large. As such, exploration with other architectures may be beneficial. While there is comparably little research on mask compression, there are still some noteworthy alternative approaches. \cite{vangansbeke2024simple_ldm_approach_for_panoptic_segm} found success in mask compression using a shallow architecture for 2D panoptic segmentation in combination with a bit-encoding of the mask. They also point out RGB-encoding as a viable option.

We show that conditioning, particularly in combination with guidance, can be an effective way of controlling the ventricular volume of generated images. While still subject to considerable variance that does not guarantee precise control over individual images, the distribution of synthetic data respected the parameters we chose at generation time. As guidance increased and conditioning values exceeded the training interval, the models exhibited an increased tendency of degeneration that rapidly reduced realism. When generated with parameters only slightly outside of distribution, synthetic images were judged by an expert neuroradiologist to be realistic (although not on the level of real images) both from a general perspective and from the perspective of displacement caused by enlarged ventricles. Nevertheless, since optimization of image quality was not a focus of this work, we believe there is significant potential for improvements here.

The results of the downstream task indicated that the synthetic data does indeed help the segmentation models perform better on cases of enlarged ventricles. The models trained with synthetic data were notably more consistent than \NNUreal, even more so than SynthSeg. In binary segmentation, the synthetic variants both outperformed SynthSeg in ventricular volume prediction error and \NNUaug{} achieved better Dice and IoU scores, while being trained on far less data. As \NNUreal{} was doubtlessly trained on more realistic images, we mainly attribute this to the more even distribution of ventricular volumes. However, to determine the precise reason for this may require further ablation experiments. Finally, we also point out that our test sets are on the smaller side, and having a much larger dataset would provide a considerably more precise and robust estimate of the true performance gap in the models. 

Limitations of this work include that we have spent limited resources on optimizing the diffusion model part of the mask generator. Given its role in generating a latent representation to be decoded by the autoencoder, it likely plays a significant role in the diversity of masks being generated. While we attempted to promote diversity by training for a shorter duration, a proper analysis on the success of this approach for masks remains the subject of future work. Another obvious improvement of this work that needs attention in the future involves the scheduler. In this work, we used the original linear scaling from \cite{DDPM}, but there are compelling arguments (\cite{chang2023designFundamentalsSurvey}) for the use of other schedules like cosine-based ones, as well as other samplers for generation. Furthermore, given the propensity of diffusion models to replicate training data (\cite{dar2024unconditional_memorize_patient_data}) and their well-documented lack of diversity compared to real data in a natural image context (\cite{marwood2023diversity_diffusion}), the quality of the synthetic dataset as a whole may still offer room for improvement. While we found that individual images appeared distinct and diverse enough according to an expert evaluation, the only indication we have for the quality of the generated synthetic datasets as a whole are their performance on downstream tasks, which may not be representative. Lastly, we point out some general limitations of our models. Working with 3D medical data is challenging due to the cubic scaling of the data. We largely circumvented this issue by universally downscaling the data and using an LDM approach. Depending on the application, however, higher resolution data with more detail may be necessary. Other work have attempted to tackle this issue by separating the mask generation task into a sequential generation of slices conditioned on previous slices (\cite{han2023medgen3d}), though this involves a significantly more complicated architecture. Another possibility may be the use of a separate super-resolution model in order to upsample low-resolution synthetic images.

The success of synthetic data in this current work raises some interesting questions. Generally, purely synthetic data is known to perform considerably worse than real data. While \cite{Fernandez_2022_trained_with_fully_synthetic_data} found comparative success in similar segmentation tasks with synthetic data, its performance still falls short of real data, which is only echoed by an increasing volume of research from the natural image domain (\cite{azizi2023synthetic_improves_imagnet,marwood2023diversity_diffusion}). While we have speculated about potential reasons for this, we believe this work offers a compelling starting point for future work to further investigate the success of purely synthetic data.

%\section{Conclusion}

In conclusion, we have proposed a framework for generating synthetic data and labels for brain MRI with varying ventricular volumes. As a proof-of-concept, we demonstrated its effectiveness in augmenting a dataset of real patients, particularly those with few or no examples of large ventricles. Additionally, we showed that incorporating these synthetically generated images and labels during training improved performance in the downstream segmentation task tested on a data set of patients with abnormally enlarged ventricles.

%In the context of ventricular segmentation, this work serves as a proof of concept for the potential of synthetic data to not only serve as a stand-in for real data, but to improve model performance. To achieve this, we configured the distribution of the generated synthetic data to align better with the known disease symptom of enlarged ventricles in the targeted cohort. The use of conditioning and guidance with LDMs proved instrumental in controlling the ventricular volume of the generated synthetic data. Image quality was judged by an expert neuroradiologist to be realistic, albeit not on the level of real data. Even so, the segmentation models trained on the synthetic data still managed to outperform state-of-the-art model SynthSeg in ventricular segmentation. We believe this work offers fertile ground for further investigation into the potential of synthetic data in a medical context.

%%%%%%%%%%%%%%%%%%%%%%%%%%%%%%%%%%%%%%%%%%%%%%%%%%%%%%%%%%%%%%%%%%%%%%%
% Mandatory Sections. Please complete, especially for final publication
%%%%%%%%%%%%%%%%%%%%%%%%%%%%%%%%%%%%%%%%%%%%%%%%%%%%%%%%%%%%%%%%%%%%%%%

% Acknowledgements.
% Please include any funding, intellectual contributions not included in the authorship, and any other acknowledgements.
\acks{This research received no external funding.}

% Ethical Standards.
% Please edit with the appropriate ethics considerations for your work. Include any pertinent IRB information, etc.
%
% Please note that the submission requirements included:
% The work presented must follow appropriate ethical standards in conducting research and writing the manuscript, following all applicable laws and regulations regarding treatment of animals or human subjects.
\ethics{The work follows appropriate ethical standards in conducting research and writing the manuscript, following all applicable laws and regulations regarding treatment of animals or human subjects.}

% Conflict of Interest
% Declaration of possible conflicts of interest: Authors must disclose any financial, organisational, commercial or personal conflicts of interest that might bias their work.
% If no conflicts, please say "We declare we don't have conflicts of interest."
\coi{We declare we don't have conflicts of interest.}

% Data availability
\data{Data supporting reported results can be obtained via contact with the corresponding author upon reasonable request and legal approval. The data are not publicly available due to no public data sharing agreement.}

\bibliography{main}

% needed in general before appendices.
% \newpage

% Appendix is optional
\clearpage
\appendix
\counterwithin{figure}{section}
\section{Architecture}

\begin{figure}[ht]
    \centering
    \includegraphics[width=\linewidth]{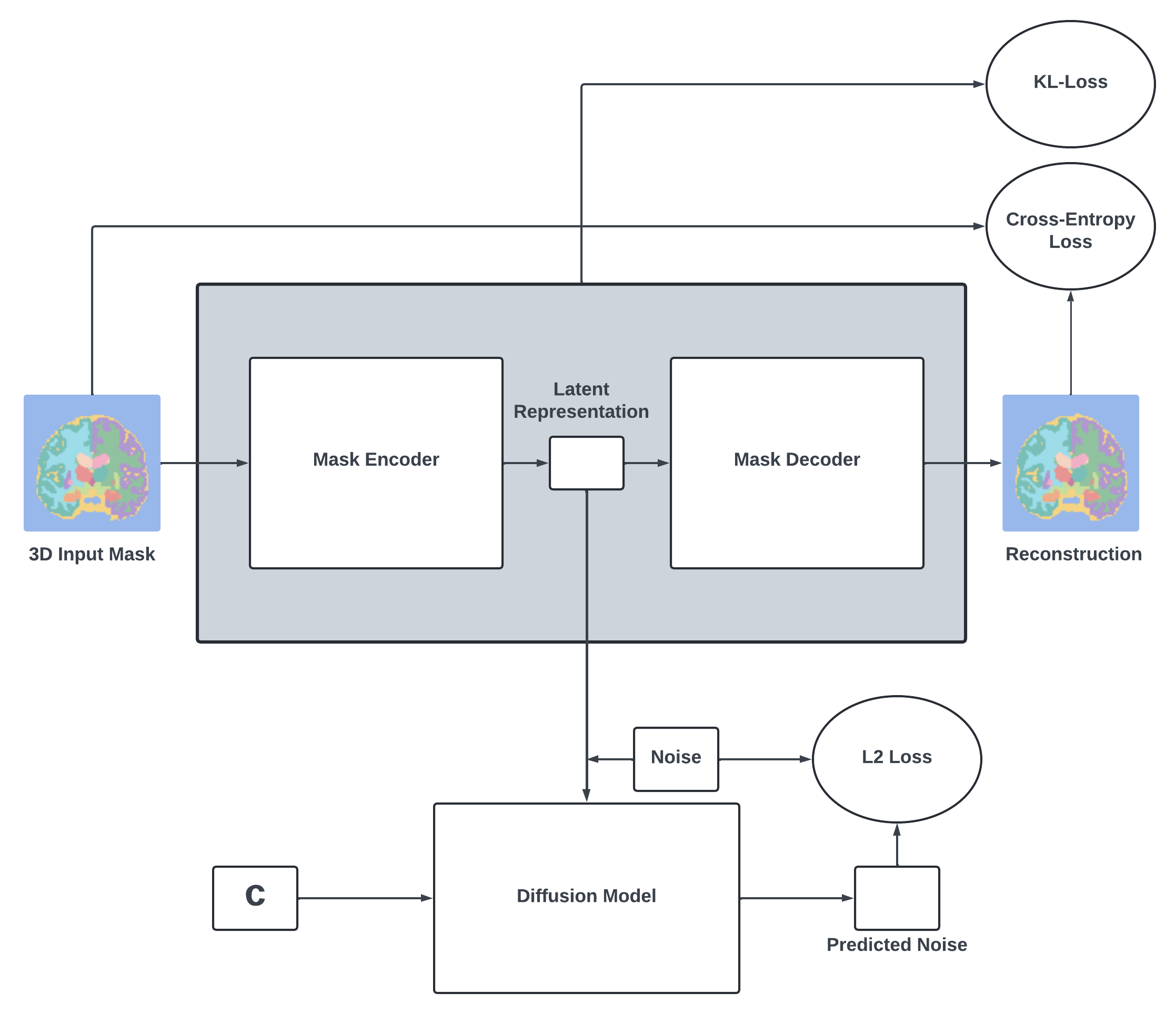}
    \caption{Architecture overview for training of the mask generator. While both autoencoder and diffusion model are presented, the autoencoder is trained separately and then used statically for training the diffusion model. Note there is a slight simplification in the L1 loss of the diffusion model. In reality, the L1 loss is calculated between the predicted noise between the input mask and the mask with added noise, and the true noise that was actually added.}
    \label{fig:mask_generator_overview}
\end{figure}

\begin{figure}[t!]
    \centering
    \includegraphics[width=\linewidth]{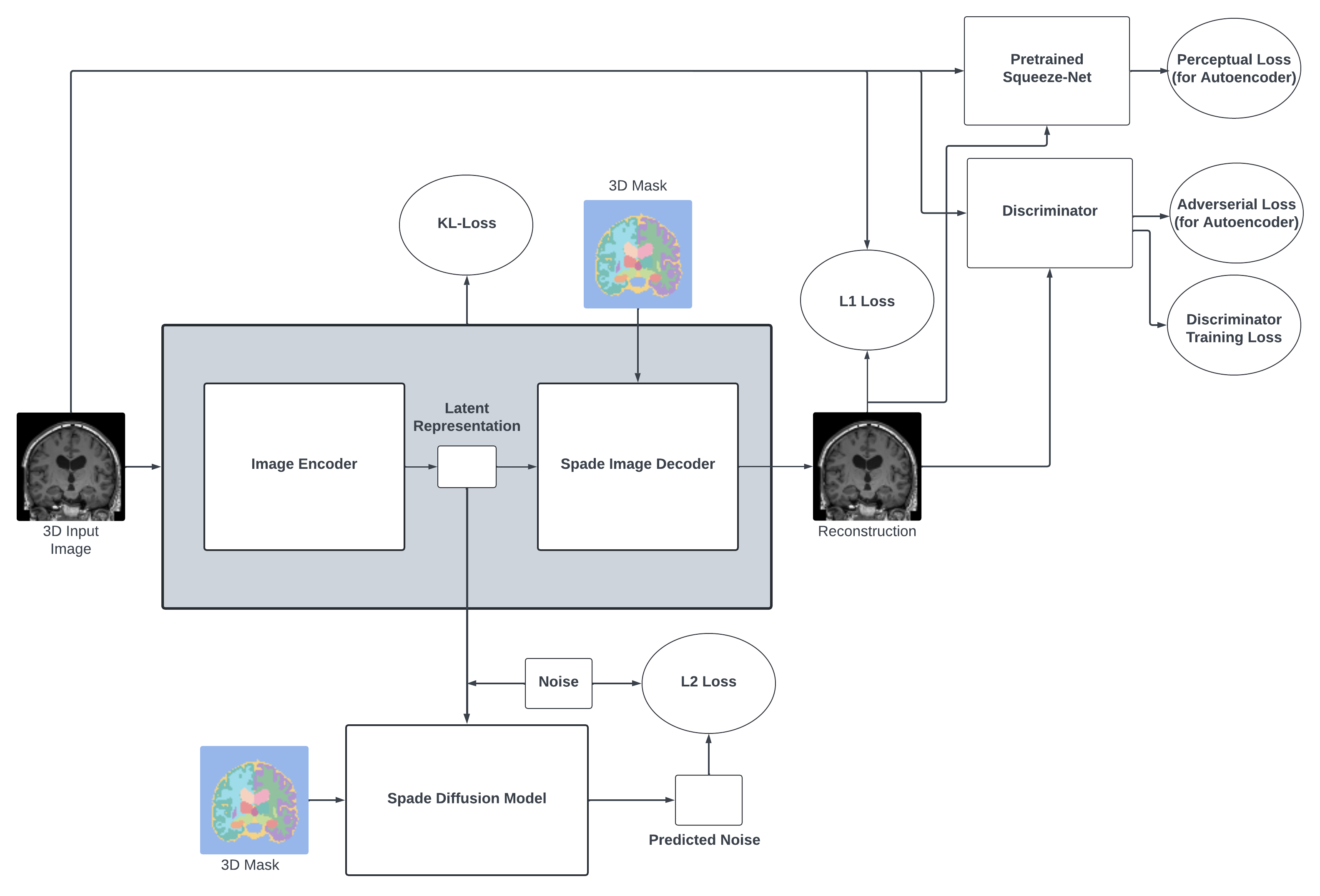}
    \caption{Architecture overview for training of the Spade Model. While both autoencoder and diffusion model are presented, the autoencoder is trained separately and then used statically for training the diffusion model. Note there is a slight simplification in the L1 loss of the diffusion model. In reality, the L1 loss is calculated between the predicted noise between the input image and the image with added noise, and the true noise that was actually added.}
    \label{fig:Spade_model_architecture}
\end{figure}

\pagebreak

\section{Experiments}

\begin{figure}[t!]
    \centering
    \includegraphics[width=\linewidth]{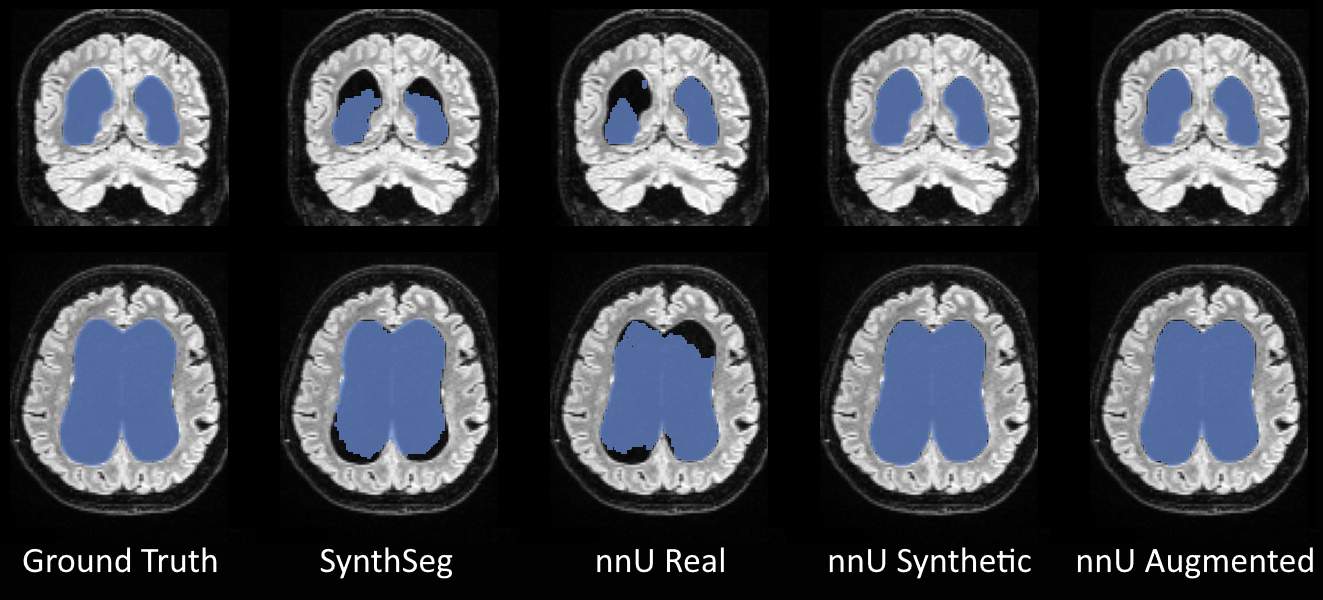}
    \caption{Slices overlaid with masks from patient 42 of the test set (the patient with the largest ventricles; see \autoref{fig:data_dist_prediction}) for the different models. While SynthSeg and \NNUreal{} considerably underestimate the true size of the ventricles, the two models trained on customized synthetic data (\NNUsyn{} and \NNUaug) segment the ventricles much better.}
    \label{fig:side_by_side_patient_42}
\end{figure}

\end{document}